\documentclass[11pt,a4paper]{article}
\usepackage[hyperref]{acl2021}
\usepackage{times}
\usepackage{latexsym}

\usepackage{microtype}

\aclfinalcopy %
\def\aclpaperid{3056} %

\usepackage{subcaption}
\usepackage{comment}
\usepackage{booktabs}
\usepackage{multicol}
\usepackage{multirow}
\usepackage{rotating}
\usepackage{longtable}

\usepackage{siunitx}
\makeatletter
\renewcommand{\paragraph}{%
\@startsection{paragraph}{4}%
{\z@}{1.0ex \@plus 1ex \@minus .2ex}{-1em}%
{\normalfont\normalsize\bfseries}%
}
\makeatother

\usepackage{tikz}
\usetikzlibrary{backgrounds,calc,shadings,shapes.arrows,arrows,shapes.symbols,shadows,positioning,decorations.markings,backgrounds,arrows.meta}
\usepackage{pgfplots}
\DeclareUnicodeCharacter{2212}{−}
\usepgfplotslibrary{groupplots,dateplot}
\usetikzlibrary{patterns,shapes.arrows}
\pgfplotsset{compat=newest}

\definecolor{MyBlue}{rgb}{0.0, 0.18, 0.65}
\colorlet{HighlightColor}{MyBlue}
\definecolor{deepskyblue}{rgb}{0.0, 0.75, 1.0}

\usepackage{xspace}
\makeatletter
\DeclareRobustCommand\onedot{\futurelet\@let@token\@onedot}
\def\@onedot{\ifx\@let@token.\else.\null\fi\xspace}

\def\eg{e.g\onedot} 
\def\ie{i.e\onedot} 
 
 \def\vs{vs\onedot}

\makeatother

\newcommand{\textsup}[1]{{\normalfont\normalsize\textsuperscript{#1}}} 
\newcommand{\var}[1]{\texttt{#1}}
\newcommand{\highlight}[1]{\textbf{\underline{#1}}}

\usepackage{amsmath}
\usepackage{amssymb}
\usepackage{amsthm}
\usepackage{bbm}
\usepackage{bm}

\usepackage{xspace}
\makeatletter
\DeclareRobustCommand\onedot{\futurelet\@let@token\@onedot}
\def\@onedot{\ifx\@let@token.\else.\null\fi\xspace}
\def\eg{e.g\onedot} 
\def\ie{i.e\onedot} 
 
 \def\vs{vs\onedot}

\makeatother

\theoremstyle{plain}

\newtheorem*{lemma*}{Lemma}

\newtheorem*{subproblem*}{Subproblem}

\newtheorem*{problem*}{Problem}

\newtheorem*{conjecture*}{Conjecture}

\theoremstyle{definition}

\newtheorem*{definition*}{Definition}

\makeatletter
\newcommand*\dotp{\mathpalette\dotp@{.5}}
\newcommand*\dotp@[2]{\mathbin{\vcenter{\hbox{\scalebox{#2}{$\m@th#1\bullet$}}}}}
\makeatother

\usepackage{xparse}
\makeatletter
\DeclareDocumentCommand{\newmathcommand}{ m O{0} m }{%
	\ifcsname\expandafter\@gobble\string#1\space\endcsname
	\expandafter\expandafter\expandafter\let\expandafter\csname old\string#1\expandafter\endcsname\expandafter=\csname\expandafter\@gobble\string#1\space\endcsname
	\else
	\expandafter\let\csname old\string#1\endcsname=#1
	\fi
	\expandafter\newcommand\csname new\string#1\endcsname[#2]{#3}
	\DeclareRobustCommand#1{%
		\ifmmode
		\expandafter\let\expandafter\next\csname new\string#1\endcsname
		\else
		\expandafter\let\expandafter\next\csname old\string#1\endcsname
		\fi
		\next
	}%
}
\makeatother

\newmathcommand{\a}{{\mathbf{a}}}
\newmathcommand{\b}{{\mathbf{b}}}

\newmathcommand{\c}{{\mathbf{c}}} 

\newmathcommand{\d}{{\mathbf{d}}}

\newmathcommand{\k}{{\mathbf{k}}}
\newmathcommand{\m}{{\mathbf{m}}}
\newmathcommand{\o}{{\mathbf{o}}}
\newmathcommand{\p}{{\mathbf{p}}}
\newmathcommand{\q}{{\mathbf{q}}}

\newmathcommand{\t}{{\mathbf{t}}}

\newmathcommand{\u}{{\mathbf{u}}}
\newmathcommand{\v}{{\mathbf{v}}}
\newcommand{\x}{\mathbf{x}}
\newcommand{\y}{{\mathbf{y}}}

\newmathcommand{\G}{{\mathbf{G}}}
\newmathcommand{\H}{{\mathbf{H}}}
\newmathcommand{\M}{{\mathbf{M}}}

\newmathcommand{\P}{{\mathbf{P}}}
\newmathcommand{\Q}{{\mathbf{Q}}}
\newmathcommand{\S}{{\mathbf{S}}}

\title{Lightweight Adapter Tuning for Multilingual Speech Translation}

\author{Hang Le\textsup{1}\qquad Juan Pino\textsup{2} \qquad Changhan Wang\textsup{2}\\
\bfseries\large
Jiatao Gu\textsup{2}\qquad Didier Schwab\textsup{1} \qquad Laurent Besacier\textsup{1,3}\\
{\textsuperscript{1}Univ. Grenoble Alpes, CNRS, LIG \qquad \textsuperscript{2}Facebook AI \qquad
\textsuperscript{3}Naver Labs Europe}\\\normalsize
\texttt{\{hang.le, didier.schwab, laurent.besacier\}@univ-grenoble-alpes.fr}\\\normalsize
\texttt{\{juancarabina, changhan, jgu\}@fb.com}
  }

\date{}

\begin{document}
\maketitle

\begin{abstract}
	{Adapter modules were recently introduced as an efficient alternative to fine-tuning in NLP. Adapter tuning consists in freezing pre-trained parameters of a model and injecting lightweight modules between layers, resulting in the addition of only a small number of task-specific trainable parameters. While adapter tuning was investigated for multilingual neural machine translation, this paper proposes a comprehensive analysis of adapters for multilingual speech translation (ST). Starting from different pre-trained models (a multilingual ST trained on parallel data or a multilingual BART (mBART) trained on non-parallel multilingual data), we show that adapters can be used to: (a) efficiently specialize ST to specific language pairs with a low extra cost in terms of parameters, and (b) transfer from an automatic speech recognition (ASR) task and an mBART pre-trained model to a multilingual ST task. {Experiments show that adapter tuning offer competitive results to full fine-tuning, while being much more parameter-efficient.}
	}
\end{abstract}

\section{Introduction}

The question of \emph{versatility} versus \emph{specialization} is often raised in the design of any multilingual translation system: is it possible to have a single model that can translate from any source language to any target one, or does it have to be multiple models each of which is in charge of one language pair?
The former is referred to as a \emph{multilingual} model, while the latter are \emph{bilingual} ones. These two paradigms have their own strengths and limitations. From a practical point of view, a multilingual model seems to be highly desirable due to its simplicity in \emph{training} and \emph{deployment}, in terms of both time and space complexities. However, in terms of 
\emph{accuracy}, a multilingual model could be outperformed by its bilingual counterparts, especially on high-resource language pairs.
 
\begin{figure}[!htb]
	\centering
    \begin{subfigure}[b]{0.6\linewidth}
		\centering
		\includegraphics[width=\linewidth]{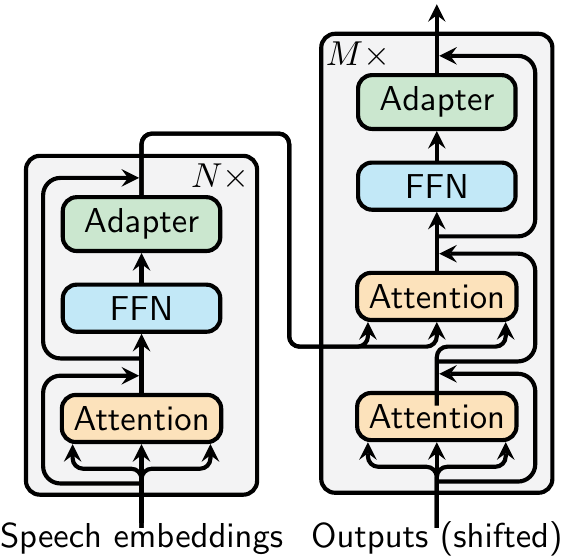}
	\caption{\label{fig:transformer}Transformer with adapters.}
	\end{subfigure}
    \begin{subfigure}[b]{0.38\linewidth}
		\centering
		\includegraphics[width=0.7\linewidth]{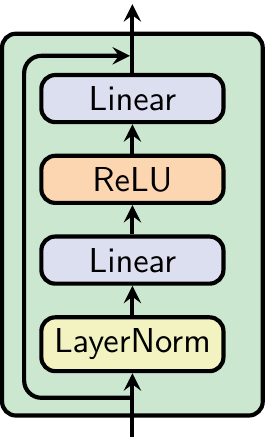}
		\caption{\label{fig:adapter_cell} An adapter cell.}
	\end{subfigure}%
	\caption{\label{fig:adapters} (a) Transformer with adapters at its FFN sub-layers. For simplicity, layer normalization \cite{ba2016layer} is omitted. During fine-tuning, only the adapters are trained. (b) A typical adapter architecture.}
	\vspace{-6mm}
\end{figure}
 
In practice, a certain trade-off between the aforementioned factors (and thus more generally between versatility and specialization) has often to be made, and depending on the application, one can be favored more than the other. 
One way to move along the spectrum between multilingual and bilingual models is to use adapter tuning which consists in freezing pre-trained parameters of a multilingual model and injecting lightweight modules between layers resulting in the addition of a small number of language-specific trainable parameters. While adapter tuning was investigated for multilingual neural machine translation (NMT)~\cite{bapna2019simple}, to our knowledge, this paper proposes the first comprehensive analysis of adapters for multilingual speech translation.

Our contributions are the following: (1) we show that both versatility and specialization can be achieved by tuning language-specific adapter modules on top of a multilingual system. Bilingual models with higher accuracy than the original multilingual model are obtained, yet keeping a low maintenance complexity;
(2) starting from a different initialization point, we show that adapters can also be used as a glue to connect off-the-shelf systems (an automatic speech recognition (ASR) model and a multilingual denoising auto-encoder mBART~\cite{liu2020multilingual,tang2020multilingual}) to perform the multilingual ST task. Extensive experiments on the MuST-C dataset~\cite{di2019must} show that adapter-based fine-tuning can achieve very competitive results to full fine-tuning---while being much more parameter-efficient---in both standard and low-resource settings. Our code based on \textsc{fairseq S2T}~\cite{wang2020fairseq} is publicly available.\footnote{\url{https://github.com/formiel/fairseq/tree/master/examples/speech_to_text/docs/adapters.md}} %

\section{Related Work}

Adapter layers (or \textit {adapters} for short) were first proposed in computer vision~\cite{rebuffi2017learning}, then explored for text classification tasks in NLP~\cite{houlsby2019parameter}. Adapters are generally inserted %
between the layers of a pre-trained network and finetuned on the adaptation corpus. \newcite{bapna2019simple} studied adapters in the context of NMT and evaluated them on two tasks: domain adaptation and massively multilingual NMT. \newcite{philip2020language} later introduced monolingual adapters for zero-shot NMT. Other research groups made contributions on the use of adapters in NLP~\cite{pfeiffer2020mad,pfeiffer2020adapterfusion} and a framework built on top of HuggingFace Transformers library~\cite{wolf2019transformers} was also released to facilitate the downloading, sharing, and adapting state-of-the-art pre-trained models with adapter modules \cite{pfeiffer2020adapterhub}.
Also very relevant to our paper is the work of~\newcite{stickland2020recipes} where adapters are used to adapt pre-trained BART~\cite{lewis2020bart} and mBART25 (multilingual BART pre-trained on 25 languages)~\cite{liu2020multilingual} to machine translation.   %

As far as speech processing is concerned, adapters were mostly used in ASR
\cite{kannan2019large,lee2020adaptable,winata2020adapt,zhu2020multilingual}. 
Recently, they have also been explored for ST as well but in a limited scope. \newcite{escolano2020enabling} addressed a very specific setting (zero-shot ST), while \newcite{li2020multilingual} and \newcite{gallego2021upc} used only one or two adapters right after a Transformer encoder.%

\section{Adapters for Speech Translation}

In this section, we describe the integration of adapters into a given backbone model for speech translation. As the Transformer~\cite{vaswani2017attention} has become common in speech processing,\footnote{For speech applications~\cite{inaguma2020espnet,wang2020fairseq}, the embedding layer of the encoder is often a small convolutional neural network \cite{fukushima1982neocognitron,lecun1989backpropagation}.} it will be used as our backbone. %
Our method, however, can be applied to any other architectures, \eg, dual-decoder Transformer~\cite{le2020dualdecoder}. %

Adapter modules can be introduced into a Transformer in a \emph{serial} or \emph{parallel} fashion. Consider a layer represented by a function $f$ that produces an output $\y$ from an input $\x$, \ie, $\y = f(\x)$. This can be an entire encoder or decoder layer, or just one of their sub-layers (\eg, the self-attention or the final feed-forward network (FFN) component). Suppose that our adapter layer is represented by a function $g$. The new ``adapted'' output is then given by:
\begin{equation*}
    \y_{\mathrm{serial}} = g(f(\x)),\quad
    \y_{\mathrm{parallel}} = f(\x) + g(\x).
\end{equation*}
Intuitively, a serial adapter modifies the output directly, while a parallel one performs the operations in parallel before merging its output to the layer. In Figure~\ref{fig:transformer}, we show an example of serial adapters being integrated to the Transformer, or more precisely to its FFN sub-layers. A common adapter module \cite{bapna2019simple} is presented in Figure~\ref{fig:adapter_cell}. Here $g$ is a small FFN with a residual connection. The first linear layer is typically a down projection to a bottleneck dimension, and the second one projects the output back to the initial dimension. Bottleneck allows us to limit the number of parameters. Other adapter architectures also exist, \eg, \newcite{stickland2019bert} explored parallel adapters consisting of a multi-head attention (MHA) layer in a multi-task setup.

For multilingual ST, we adopt the following general recipe for adapter-based fine-tuning. Starting from a pre-trained backbone, an adapter is added for each language pair and then finetuned on the corresponding bilingual data (while the rest of the backbone is frozen). The pre-trained backbone plays a crucial role in this recipe. We explore two common scenarios to obtain this pre-trained model, namely \emph{refinement} and \emph{transfer learning}. We present them in details, together with extensive experimental results, in Section \ref{sec:adapting_finetuning} and \ref{sec:transfer}. In the next section, we present our experimental setup.

\section{Experimental Setup}
\begin{table*}[htb!]
	\centering
	\resizebox{\linewidth}{!}{
		\begin{tabular}{c|c| c | ccc | cc | l | llll llll | c }
			\toprule
			& & & \multicolumn{3}{c|}{\textbf{Adapter}} & \multicolumn{2}{c|}{\textbf{Finetune}} & {\textbf{\# params (M)}} & & & & & & & & & \\
			
			& \textbf{Dict} & $D$ &$d$ & \textbf{ENC} & \textbf{DEC}&
			\textbf{ENC}& \textbf{DEC} &
			\textbf{trainable/total} & 
			\textbf{de} & \textbf{es} & \textbf{fr} & \textbf{it} & 
			\textbf{nl} & \textbf{pt} & \textbf{ro} & \textbf{ru} & \textbf{avg}  \\
			\midrule
			
			\multicolumn{9}{r|}{Training data (hours)}& 408 & 504 & 492 & 465 & 442 & 385 & 432 & 489 & \\
			\midrule
			1 & {mono} & {\multirow{8}{*}{\rotatebox[origin=c]{90}{256}}} & - & - & - & - & - & 8$\times$31.1/8$\times$31.1& 22.16 & 30.42& 27.92& 22.92& 24.10& 27.19 & 21.51& 14.36 & 23.82\\
			
			2 & {multi} & & - &- & - & - & - & 32.1/32.1 &  22.37 & 30.40 & 27.49 & 22.79 & 24.42 & 27.32 & 20.78 & 14.54 & 23.76\\
			
			\cmidrule{4-18}
			3 & {multi} & & 64 &- & \checkmark & - & - & 8$\times$0.2/33.7& 22.32 & 30.50 & 27.55 & 22.91 & 24.51 & 27.36 & 21.09 & 14.74 & 23.87\\
			
			4 & {multi} &  & 64 & \checkmark & \checkmark & - & - & 8$\times$0.6/36.9 & 22.75 & 31.07 & 28.03 & 23.04 & 24.75 & 28.06 & 21.20 & 14.75 & 24.21\\
			
			\cmidrule{4-18}
			5 & {multi} &  & 128 &- & \checkmark & - & - & 8$\times$0.4/35.3& 22.45 & 30.85 & 27.71 & 23.06 & 24.57 & 27.52 & 20.93 & 14.57 & 23.96\\
			
			6 & {multi} &  & 128 & \checkmark & \checkmark & - & - & 8$\times$1.2/41.7 & 22.84$^{*}$ & 31.25$^{*}$ & 28.29$^{*}$ & 23.27$^{*}$ & 24.98$^{*}$ & 28.16$^{*}$ & 21.36$^{*}$ & 14.71 & 24.36\\
			
			\cmidrule{4-18}
			7 & {multi} &  & - &- & - & - & \checkmark & 8$\times$14.6/8$\times$32.1& \underline{23.49} & 31.29 & {28.40} & 23.63 & 25.51 & 28.71 & 21.73 & 15.22 & 24.75\\
			
			8 & {multi} &  & - &- & - & \checkmark & \checkmark & 8$\times$32.1/8$\times$32.1& 23.13$^{*}$ & \underline{31.39}$^{*}$ & \underline{28.67}$^{*}$ & \underline{23.80}$^{*}$ & \underline{25.52}$^{*}$ & \underline{29.03}$^{*}$ & \underline{22.25}$^{*}$ & \underline{15.44}$^{*}$ & \underline{24.90}\\
			\midrule
			
			9 & {mono} & {\multirow{10}{*}{\rotatebox[origin=c]{90}{512}}} & - & - & - & - & - & 8$\times$74.3/8$\times$74.3& 21.93 & 30.46 & 27.90 & 22.64 & 23.98 & 25.98 & 20.50 & 14.01 & 23.42 \\
			
			10 & {multi} & & - &- & - & - & - & 76.3/76.3 & 23.98 & 32.47 & 29.24 & 24.97 & 26.20 & 29.81 & 22.74 & 15.30 & 25.59 \\
			
			\cmidrule{4-18}
			11 & {multi} &  & 64 &-  & \checkmark & - & - & 8$\times$0.4/79.5& 24.24 & 32.52 & 29.47 & \color{HighlightColor}24.74 & \color{HighlightColor}26.13 & \color{HighlightColor}29.72 & \color{HighlightColor}22.53 & \color{HighlightColor}15.25 & \color{HighlightColor}25.57 \\
			
			12 & {multi} &  & 64 & \checkmark & \checkmark & - & - & 8$\times$1.2/85.9 & 24.13 & 32.80 & 29.55 & \color{HighlightColor}24.90 & \color{HighlightColor}26.04 & 30.25 & \color{HighlightColor}22.73 & 15.31 & 25.72 \\
			
			\cmidrule{4-18}
			13 & {multi} &  & 128 &- & \checkmark & - & - & 8$\times$0.8/82.7& 24.34 & 32.86 & 29.51 & \color{HighlightColor}24.73 & \color{HighlightColor}26.15 & 30.01 & \color{HighlightColor}22.58 & \color{HighlightColor}15.07 & 25.66\\
			
			14 & {multi} &  & 128 &\checkmark & \checkmark & - & - & 8$\times$2.4/95.5& 24.30 & 32.61 & 29.72$^{*}$ & 25.07 & 26.29 & 30.46$^{*}$ & 22.99 & 15.47 & 25.86\\
			
			\cmidrule{4-18}
			15 & {multi} & & 256 &- & \checkmark & - & - & 8$\times$1.6/89.1& 24.38 & 32.78 & 29.69 & \color{HighlightColor}24.72 & 26.25 & 29.93 & \color{HighlightColor}22.63 & 15.40 & 25.72\\
			
			16 &{multi} & & 256 &\checkmark & \checkmark & - & - & 8$\times$4.8/114.7 & 24.61 & 32.94 & 29.67 & \highlight{25.12} & \color{HighlightColor}26.16 & \highlight{30.53} & \color{HighlightColor}22.66 & 15.31 & 25.88\\
			
			\cmidrule{4-18}
			17 & {multi} & & - & - &-& - & \checkmark & 8$\times$35.5/8$\times$36.3& \highlight{24.67} & \highlight{33.12} & \highlight{30.11} & {25.05} & \highlight{26.33} & 29.85 & \highlight{23.04} & \highlight{15.61} & \highlight{25.97}\\
			
			18 & {multi} &  & - & - & - & \checkmark & \checkmark & 8$\times$76.3/8$\times$76.3& 24.54$^{*}$ & 32.95$^{*}$ & 29.96$^{*}$ & 25.01 & 26.31 & {30.04} & \color{HighlightColor}22.66 & 15.54$^{*}$ & 25.88\\
			\bottomrule
		\end{tabular}
	}
	\caption{\label{tab:adating_finetuning_ablation}BLEU on MuST-C dev set for \textbf{refinement}. In the \textbf{Dict} column, \var{mono} and \var{multi} mean, respectively, monolingual and multilingual dictionary. $D$ is the Transformer hidden dimension. In the \textbf{Adapter} group, $d$ is the adapter bottleneck dimension, \var{ENC} and \var{DEC} mean adding adapters to encoder and decoder, respectively; and idem for the \textbf{Finetune} group. Rows 1--2 and rows 9--10 represent our bilingual and multilingual baselines for each $D$. Values lower than the multilingual baselines are colored in {\color{HighlightColor}blue}. The highest values in each group of $D$ are \underline{underlined}, while the highest values of each column are in \textbf{bold} face. {Furthermore, we select the top configurations (6, 8, 14, 18) and perform statistical significance test using bootstrap re-sampling~\cite{koehn2004statistical}. Results passing the test (compared to the corresponding multilingual baselines, with $p$-value $< 0.05$) are marked with a star.}}
	\vspace{-3mm}
\end{table*}

\subsection{Dataset}

\paragraph{MuST-C}
We evaluate our recipes on MuST-C~\cite{di2019must}, a large-scale one-to-many ST dataset from English to eight target languages including Dutch (\texttt{nl}), French (\texttt{fr}), German (\texttt{de}), Italian (\texttt{it}), Portuguese (\texttt{pt}), Romanian (\texttt{ro}), Russian (\texttt{ru}), and Spanish (\texttt{es}). Each direction includes a triplet of speech, transcription, and translation. Sizes range from 385 hours (\texttt{pt}) to 504 hours (\texttt{es}). %

\paragraph{MuST-C-Imbalanced}
{We built a low-resource version of MuST-C, called MuST-C-Imbalanced, in which we randomly keep only $X$\% of the original training data, where $X = 100$ for \texttt{es}, \texttt{fr}; $X = 50$ for \texttt{ru}, \texttt{it}; $X = 20$ for \texttt{nl}, \texttt{ro}; and $X = 10$ for \texttt{de}, \texttt{pt} (same order of the languages in the original MuST-C if we sort them in decreasing amount of data). The amount of speech data ranges from 41 hours (\texttt{de}) to 504 hours (\texttt{es}) in this version, better reflecting real-world data imbalance scenarios.}

\subsection{Implementation details}
Our implementation is based on the \textsc{fairseq S2T} toolkit~\cite{wang2020fairseq}. We experiment with two architectures: a small Transformer model with dimension $D=256$ and a medium one where $D=512$. All experiments use the same encoder with 12 layers. The decoder has 6 layers, except for the transfer learning scenario where we used the mBART decoder for initialization. We used 8k and 10k unigram vocabulary~\cite{kudo2018sentencepiece} for bilingual and multilingual models, respectively.
The speech features are 80-dimensional log mel filter-bank. Utterances having more than 3000 frames are removed for GPU efficiency. We used SpecAugment~\cite{park2019specaugment} with LibriSpeech basic (LB) policy for data augmentation.

We used the Adam optimizer~\cite{kingma2015adam} with learning rate linearly increased for the first 10K steps to a value $\eta_{\max}$, then decreased proportionally to the inverse square root of the step counter. For all adapter experiments, $\eta_{\max}$ is set to \num{2e-3}. For the others, however, we perform a grid search over three values $\{\num{2e-3}, \num{2e-4}, \num{2e-5}\}$ and select the best one on the dev set, as they are more sensitive to the learning rate.%

\section{Refinement}
\label{sec:adapting_finetuning}

\begin{table*}[htb!]
	\centering
	\resizebox{\linewidth}{!}{
		\begin{tabular}{c| c | ccc | cc | l | cccc cccc | c }
			\toprule
			& & \multicolumn{3}{c|}{\textbf{Adapter}} & \multicolumn{2}{c|}{\textbf{Finetune}} & {\textbf{\# params (M)}} & & & & & & & & & \\
			
			 & $D$ &$d$ & \textbf{ENC} & \textbf{DEC}&
			\textbf{ENC}& \textbf{DEC} &
			\textbf{trainable/total} & 
			\textbf{de} & \textbf{es} & \textbf{fr} & \textbf{it} & 
			\textbf{nl} & \textbf{pt} & \textbf{ro} & \textbf{ru} & \textbf{avg}  \\
			\midrule
			
			\multicolumn{8}{r|}{Training data (hours)}& 41 & 504 & 492 & 232 & 89 & 38 & 86 & 245 & \\
			\midrule

			1 & {\multirow{3}{*}{\rotatebox[origin=c]{90}{256}}} & - &- & - & - & - & 32.1/32.1 & 15.99 & 30.51  & 28.17 & 21.80 & 20.27 & 22.47 & 17.38 & 13.18 & 21.22\\

			2 &  & 128 & \checkmark & \checkmark & - & - & 8$\times$1.2/41.7 & \underline{17.02} & 30.71 & \underline{28.42} & 22.37 & \underline{21.01} & \underline{23.74} & \underline{18.55} & 13.52 & \underline{21.92}\\

			3 &  & - &- & - & \checkmark & \checkmark & 8$\times$32.1/8$\times$32.1& 16.93 & \underline{30.86} & 28.34 & \underline{22.42} & 20.86 & 23.44 & 18.49 & \underline{13.63} & 21.87 \\
			\midrule

			4 & {\multirow{3}{*}{\rotatebox[origin=c]{90}{512}}} & - &- & - & - & - & 76.3/76.3 & 17.05 & 31.92 & 29.06 & 22.91 & 21.64 & 24.15 & 19.18 & 14.09 & 22.50 \\

			5 & & 256 &\checkmark & \checkmark & - & - & 8$\times$4.8/114.7 & 17.46 & \textbf{31.94} & 29.09 & \textbf{23.11} & 21.76 & \textbf{24.96} & \textbf{19.50} & 14.10& \textbf{22.74}\\

			6 &  & - & - & - & \checkmark & \checkmark & 8$\times$76.3/8$\times$76.3& \textbf{17.49} & 31.67 & \textbf{29.27} & 22.97 & \textbf{21.80} & 24.80 & 19.43 & \textbf{14.17} & 22.70\\
			\bottomrule
		\end{tabular}
	}
	\caption{\label{tab:mustc_imbalanced}{BLEU on MuST-C dev set for \textbf{refinement} in the low-resource scenario where the models were trained on MuST-C-Imbalanced dataset. We refer to Table~\ref{tab:adating_finetuning_ablation} for other notation.}}
    \vspace{-3mm}
\end{table*}

\begin{table*}[!htb]
    \centering
    \resizebox{0.9\linewidth}{!}{
    \begin{tabular}{l | l | l | cccc cccc | c}
        \toprule
         & {\textbf{Method}} & \begin{tabular}{@{}c@{}} \textbf{\# params (M)} \\ \textbf{trainable/total} \end{tabular} &  {\textbf{de}} & {\textbf{es}} & {\textbf{fr}} & 
         {\textbf{it}} & 
         {\textbf{nl}} & 
         {\textbf{pt}} & 
         {\textbf{ro}} &
         {\textbf{ru}} & 
         {\textbf{avg}} \\

         \midrule
         \multirow{3}{*}{\rotatebox[origin=c]{90}{Ours}} & Baseline & 76.3/76.3 & 24.18 & 28.28 & \textbf{34.98} & 24.62 & \textbf{28.80} & \textbf{31.13} & 23.22 & 15.88 & 26.39 \\
         & Best adapting & 8 $\times$ 4.8/76.3 & \textbf{24.63} & \textbf{28.73} & 34.75 & \textbf{24.96} & \textbf{28.80} & 30.96 & 23.70 & \textbf{16.36} & \textbf{26.61} \\
         & Best fine-tuning & 8 $\times$35.5/8 $\times$ 76.3 & 24.50 & 28.67 & 34.89 & 24.82 & 28.38 & 30.73 & \textbf{23.78} & 16.23 & 26.50 \\
         \midrule
         \multirow{3}{*}{\rotatebox[origin=c]{90}{\citeauthor{li2020multilingual}}} & \texttt{LNA-D} & 53.5/76.3 & 24.16 & 28.30 & 34.52 & 24.46 & 28.35 & 30.51 & 23.29 & 15.84 & 26.18 \\
         & \texttt{LNA-E} & 48.1/76.3 & 24.34 & 28.25 & 34.42 & 24.24 & 28.46 & 30.53 & 23.32 & 15.89 & 26.18 \\
         & \texttt{LNA-E,D} & 25.3/76.3 & 24.27 & 28.40 & 34.61 & 24.44 & 28.25 & 30.53 & 23.27 & 15.92 & 26.21 \\
         \bottomrule
    \end{tabular}
    }
    \caption{\label{tab:comparison_to_li_et_al}BLEU on MuST-C test set. Our method compares favorably with \cite{li2020multilingual}.}
    \vspace{-3mm}
\end{table*}

In this section, a fully trained multilingual ST backbone is further refined on each language pair to boost the performance and close potential gaps with bilingual models. We compare adapter tuning with other fine-tuning approaches as well as the bilingual and multilingual baselines (the latter being the starting point for all fine-tuning approaches)~\cite{bapna2019simple}. %
Starting from these backbones, we either add language-specific adapters and train them only, or we finetune the backbone on each language pair, either fully or partially. All these trainings are performed on MuST-C. The results are shown in Table~\ref{tab:adating_finetuning_ablation}. There are two main blocks corresponding to two architectures: $D=256$ (small) and $D=512$ (medium). Rows 1 and 9 provide the bilingual baselines, while rows 2 and 10 serve as the multilingual baselines for each block. %
{In addition, we compare adapter-tuning with full fine-tuning and multilingual-training (baseline) on MuST-C-Imbalanced. Table~\ref{tab:mustc_imbalanced} displays the results for this set of experiments.}

\paragraph{Bilingual \vs Multilingual} For the small architecture ($D=256$), the bilingual models slightly outperform their multilingual counterpart (rows 1, 2). Looking further into 
the performance of each language pair, the multilingual model is able to improve the results for 4 out of 8 pairs (\texttt{de}, \texttt{nl}, \texttt{pt}, \texttt{ru}), mainly those in the lower-resource direction, but the joint multilingual training slightly hurts the performance of higher-resource pairs such as \texttt{es}, \texttt{fr}, \texttt{it}, and \texttt{ro}. Finally, we observe that the medium model ($D=512$) performs better in the multilingual setting than the bilingual one (rows 9, 10).  %

\paragraph{Adapter tuning \vs Fine-tuning} Both 
recipes yield improvements over the multilingual baseline and recover the lost performance of higher-resource directions compared to the bilingual baseline for the small model ($D=256$). For the medium one ($D=512$), one adapter tuning (row 14) can slightly improve the scores in all directions and even approach the results of the best fine-tuning experiment (row 17) while maintaining much lower model sizes (95.5M \vs 8$\times$ 36.3M parameters).

\paragraph{Low-resource scenario}
The obtained results on small models show that adapter-tuning achieved the best performance, producing clear improvements over the baseline, especially for the low-resource languages: $+1.1$ BLEU on average on \texttt{nl}, \texttt{ro}, \texttt{de}, \texttt{pt}; $+0.3$ BLEU on average on \texttt{es}, \texttt{fr}, \texttt{ru}, \texttt{it}; which is competitive to full fine-tuning ($+0.9$ and $+0.4$ BLEU, respectively) while being more parameter-efficient as well as simpler for training and deployment (one model with adapters versus eight separate models). For larger models, however, the improvement is smaller: $+0.4$ BLEU on average on the lower-resource pairs and $+0.1$ on the higher-resource ones; while those of full fine-tuning are $+0.4$ and roughly no improvement, respectively.

\begin{table*}[htb!]
	\centering
	\resizebox{\linewidth}{!}{
		\begin{tabular}{c| ccc| c | l| cccc cccc | c}
			\toprule
			& \multicolumn{3}{c|}{\textbf{Adapter}} & \textbf{Finetune}& {\textbf{\# params (M)}}& & & & & & & & &\\
			
			&$d$ & \textbf{ENC} & \textbf{DEC}& \textbf{xattn}& \textbf{trainable/total} & \textbf{de} & \textbf{es} & \textbf{fr} & \textbf{it} & 
			\textbf{nl} & \textbf{pt} & \textbf{ro} & \textbf{ru} & \textbf{avg}  \\
			\midrule
			1 & - & -& -& -& 8$\times$31.1/8$\times$31.1& 22.16 & 30.42& 27.92& 22.92& 24.10& 27.19 & 21.51& 14.36 & 23.82\\
			\midrule
			2 & - & - &- & \checkmark&  38 / 486 & 18.41 & 25.42 & 23.46 & 18.44 & 20.87 & 20.55 & 17.19 & 11.79 & 19.52\\
			
			3 & 512 & - & \checkmark &- &  101 / 587 & 0.94 & 0.65 & 0.93 & 0.76 & 0.95 & 0.89 & 0.52 & 0.93 & 0.82\\
			4 & 512 & - & \checkmark &\checkmark & 139 / 587 & 21.98 & 29.47 & 27.05 & 22.89 & 24.06 & 26.34 & 21.0 & 14.35 & 23.39 \\
			5 & 512 & \checkmark & \checkmark& -& 152 / 638 & 11.04 & 18.62 & 16.10 & 12.37 & 13.18 & 14.29 & 10.62 & 6.95 & 12.90\\
			6 & 512 & \checkmark & \checkmark &\checkmark& 190 / 638 & 22.62 & 30.85 & 28.23 & 23.09 & 24.43 & 26.56 & 22.13 & 14.92 & 24.10 \\
			
			\bottomrule
		\end{tabular}
	}
	\caption{\label{tab:transfer_learning_all}BLEU on MuST-C dev set for \textbf{transfer learning} from pre-trained ASR and mBART models. We compare the results with the bilingual baselines (trained from scratch), shown in row 1 (which is identical to row 1 in Table~\ref{tab:adating_finetuning_ablation}). The column ``Finetune xattn'' means updating the cross-attention parameters. We refer to Table~\ref{tab:adating_finetuning_ablation} for other notation.}
	\vspace{-3mm}
\end{table*}

\paragraph{Results on test set} We select the best-performing fine-tuning recipes on the dev set (rows 16 and 17 in Table~\ref{tab:adating_finetuning_ablation}) for evaluation on the test set. For reference, we also include the multilingual baseline (row 10). Moreover, to go beyond conventional fine-tuning approaches, we also compare our recipes with a contemporary work in which only several components of the network are finetuned~\cite{li2020multilingual}. 
For a fair comparison, we did not use large pre-trained components such as wav2vec~\cite{baevski20wav2vec} or mBART~\cite{tang2020multilingual} but instead considered the same pre-trained components used in our previous experiments. Following~\cite{li2020multilingual}, we considered six variants: fine-tuning LayerNorm + Attention in the encoder (\texttt{LNA-E}), or the decoder (\texttt{LNA-D}), or both (\texttt{LNA-E,D}); each with or without the length adapter. %
We found that adding the length adapter did not help in our experiments. %
Table~\ref{tab:comparison_to_li_et_al} shows that our approach compares favorably with \cite{li2020multilingual} in terms of both performance and parameter-efficiency.%

\paragraph{Other comments}
For small models, the encoder adapters boost the performance ($0.3$--$0.4$ BLEU on average) in all directions (rows 3 and 4, 5 and 6, Table \ref{tab:adating_finetuning_ablation}), indicating that language-specific adapters can tweak the encoder representations to make them better suited for the decoder. In larger models, however, the impact of the encoder adapters is varied depending on languages and bottleneck dimensions. We also notice that increasing the bottleneck dimension slightly improves performance while remaining parameter-efficient. %
Fine-tuning remains the best option to optimize the models in most cases but leads to much larger model sizes. {The adapter-tuning approach is competitive to fine-tuning while being much more parameter-efficient. %
}

\section{Transfer Learning}
\label{sec:transfer}

In this section, we show that adapters can be used to combine available pre-trained models to perform a multilingual ST task. In particular, we initialize the encoder using a pre-trained ASR encoder (on MuST-C)\footnote{{Pre-training on ASR data and then transferring to ST is not new but rather standard. See, \eg, \citet{bansal2019pretraining}}.} provided by~\newcite{wang2020fairseq} and the decoder using mBART50, a multilingual denoising auto-encoder pre-trained on 50 languages~\cite{tang2020multilingual}. We tune language independent cross-attention and language-specific adapters on top of these backbone models (using MuST-C as well). 
The results presented in \autoref{tab:transfer_learning_all} highlight that fine-tuning cross-attention is crucial to transfer to multilingual ST (rows 3 and 5 show poor results without doing so). Adding adapters to the backbone decoder (row 4) or to both encoder and decoder (row 6) further boosts performance, demonstrating the ability of adapters to connect off-the-shelf models in a modular fashion. The best-performing model in this recipe (row 6) also outperforms  bilingual systems (row 1) despite having fewer trainable parameters (190M \vs 248M). It is also important to mention that while we experiment on 8 target languages of MuST-C corpus, the multilingual ST model of row 2 should be practically able to decode into 50 different target languages. Investigating such a zero-shot ST scenario is left for future work.

\section{Conclusion}
We have presented a study of adapters for multilingual ST and shown that language-specific adapters can enable a fully trained multilingual ST model to be further specialized in each language pair. With these adapter modules, one can efficiently obtain a single multilingual ST system that outperforms the original multilingual model as well as multiple bilingual systems while maintaining a low storage cost and simplicity in deployment. In addition, adapter modules can also be used to connect available pre-trained models such as an ASR model and a multilingual denoising auto-encoder to perform the multilingual speech-to-text translation task.

\section*{Acknowledgments}

This work was supported by a Facebook AI SRA grant, and was granted access to the HPC resources of IDRIS under the allocation 2020-AD011011695 made by GENCI. It was also done as part of the Multidisciplinary Institute in Artificial Intelligence MIAI@Grenoble-Alpes (ANR-19-P3IA-0003). We thank the anonymous reviewers for their insightful questions and feedback.

\bibliographystyle{acl_natbib}
\bibliography{refs,acl2021}

\appendix

\section{Parallel Adapters}
\label{sec:parallel_adapters}

In this section, we present our preliminary experiments in which we explore different positions of the parallel adapters: in parallel with either Transformer layers or their sub-layers. We perform experiments where the adapters are added to the decoder. The results are shown in Table~\ref{tab:parallel_adapters}. 

\begin{table}[htb!]
 \centering
 \resizebox{0.9\linewidth}{!}{
 	\begin{tabular}{c | ccc | l | c }
 		\toprule
 		& \multicolumn{3}{c|}{\textbf{Adapter}} & {\textbf{\# params (M)}} & \\
 		
 		&$d$ & $h$ & \textbf{type}& 
 		\textbf{trainable/total} & 
 		\textbf{en-de}  \\
 		\midrule
 		
 		1 & - &- & - & 32.1/32.1 & 22.37 \\
 		
 		2 & 128 &- & \texttt{ser} & 0.4/32.5& 22.45\\
 		
 		3 & 128 &4& \texttt{par-TL} & 0.8/32.9& 21.67\\
 		
 		4 & 128 &4 & \texttt{par-SA} &0.8/32.9 & 19.55 \\
 		
 		5 & 128 &4 & \texttt{par-XA} & 0.8/32.9& 19.22 \\
 		
 		\bottomrule
 	\end{tabular}
 }
 \caption{\label{tab:parallel_adapters}BLEU on dev set for \textbf{parallel \vs serial adapters}. In the ``Adapter'' block, $d$ is the adapter's dimension, $h$ is the number of heads, \var{ser} stands for {serial} adapters, and \var{par} stands for {parallel} ones. The suffixes denote the position of the parallel adapters: in parallel with the Transformer layer (\texttt{TL}), or with self-attention sub-layer (\texttt{SA}), or with cross-attention sub-layer (\texttt{XA}).}
\end{table}

Among the parallel variants, the one that performs operations in parallel with a full layer produces the best result. However, its performance still could not surpass the serial adapter (row 2) as well as the starting point (row 1).

\section{Specializing}
\label{sec:specializing}
\begin{table*}[!h]
 \centering
 \resizebox{\linewidth}{!}{
 	\begin{tabular}{c|c| c | ccc | cc | l | cccc cccc c }
 		\toprule
 		& & & \multicolumn{3}{c|}{\textbf{Adapter}} & \multicolumn{2}{c|}{\textbf{Finetune}} & {\textbf{\# params (M)}} & & & & & & & & & \\
 		
 		& \textbf{Dict} & $D$ &$d$ & \textbf{ENC} & \textbf{DEC}&
 		\textbf{ENC}& \textbf{DEC} &
 		\textbf{trainable/total} & 
 		\textbf{de} & \textbf{es} & \textbf{fr} & \textbf{it} & 
 		\textbf{nl} & \textbf{pt} & \textbf{ro} & \textbf{ru} & \textbf{avg}  \\
 		\midrule
 		
 		1 & {mono} & {\multirow{8}{*}{\rotatebox[origin=c]{90}{256}}} & - & - & - & - & - & 8$\times$31.1/8$\times$31.1& 22.16 & 30.42& 27.92& 22.92& 24.10& 27.19 & 21.51& 14.36 & 23.82\\
 		
 		2 & {multi} & & - &- & - & - & - & 32.1/32.1 &  22.37 & 30.40 & 27.49 & 22.79 & 24.42 & 27.32 & 20.78 & 14.54 & 23.76\\
 		
 		\cmidrule{4-18}
 		3 & {mono} & & 64 &- & \checkmark & - & - & 8$\times$4.3/8$\times$31.3& 23.28 & 30.95 & 28.31 & 23.25 & 24.76 & 27.84 & 21.55 & 14.60 & 24.32\\
 		
 		4 & {mono} &  & 64 & \checkmark & \checkmark & - & - & 8$\times$4.7/8$\times$31.7&  23.53 & 31.16 & 28.83 & 23.29 & 24.43 & 28.18 & 21.38 & 14.66 & 24.44\\
 		
 		\cmidrule{4-18}
 		5 & {mono} &  & 128 &- & \checkmark & - & - & 8$\times$4.5/8$\times$31.5 & 23.33 & 31.05 & 28.67 & 23.43 & 24.83 & 28.10 & 21.44 & 14.58 & 24.43\\
 		
 		6 & {mono} &  & 128 &\checkmark & \checkmark & - & - & 8$\times$5.3/8$\times$32.3 & \color{HighlightColor}22.09 & \color{HighlightColor}30.09 & 27.63 & \color{HighlightColor}22.53 & \color{HighlightColor}24.24 & \color{HighlightColor}27.09 & \color{HighlightColor}20.36 & \color{HighlightColor}14.19 & \color{HighlightColor}23.53 \\
 		
 		\cmidrule{4-18}
 		7 & {mono} &  & - &- & - & - & \checkmark & 8$\times$13.6/8$\times$31.1& 24.03 & 31.79 & 29.64 & 24.16 & 25.55 & 28.92 & 22.11 & 15.00 & 25.15\\
 		
 		8 & {mono} &  & - &- & - & \checkmark &  \checkmark & 8$\times$31.1/8$\times$31.1& 23.89 & 31.72 & 29.23 & 23.65 & 25.14 & 28.23 & 21.83 & 14.80 & 24.81\\
 		\midrule
 		
 		9 & {mono} & {\multirow{10}{*}{\rotatebox[origin=c]{90}{512}}} & - & - & - & - & - & 8$\times$74.3/8$\times$74.3& 21.93 & 30.46 & 27.90 & 22.64 & 23.98 & 25.98 & 20.5 & 14.01 & 23.42 \\
 		
 		10 & {multi} & & - &- & - & - & - & 76.3/76.3 & 23.98 & 32.47 & 29.24 & 24.97 & 26.20 & 29.81 & 22.74 & 15.30 & 25.59 \\
 		
 		\cmidrule{4-18}
 		11 & {mono} &  & 64 &-  & \checkmark & - & - & 8$\times$8.6/8$\times$74.7& \color{HighlightColor}23.85 & \color{HighlightColor}31.79 & 29.63 & \color{HighlightColor}24.26 & \color{HighlightColor}25.77 & \color{HighlightColor}28.97 & \color{HighlightColor}22.18 & \color{HighlightColor}15.02 & \color{HighlightColor}25.18\\
 		
 		12 & {mono} &  & 64 &\checkmark  & \checkmark & - & -&  8$\times$9.4/8$\times$75.5 & \color{HighlightColor}23.74 & 31.62 & 29.44 & \color{HighlightColor}24.02 & \color{HighlightColor}25.56 & \color{HighlightColor}29.23 & \color{HighlightColor}22.25 & 15.39 & \color{HighlightColor}25.16 \\
 		
 		\cmidrule{4-18}
 		13 & {mono} &  & 128 &-  & \checkmark & - & - & 8$\times$9.0/8$\times$75.1& \color{HighlightColor}23.91 & \color{HighlightColor}32.05 & 29.47 & \color{HighlightColor}24.08 & \color{HighlightColor}25.86 & \color{HighlightColor}29.28 & \color{HighlightColor}22.30 & \color{HighlightColor}15.28 & \color{HighlightColor}25.28\\
 		
 		14 & {mono} &  & 128 &\checkmark  & \checkmark & - & - & 8$\times$10.6/8$\times$76.7 & 23.98 & \color{HighlightColor}32.28 & 29.40 & \color{HighlightColor}24.46 & \color{HighlightColor}25.46 & \color{HighlightColor}29.28 & \color{HighlightColor}21.90 & \color{HighlightColor}15.15 & \color{HighlightColor}25.24\\
 		
 		\cmidrule{4-18}
 		15 & {mono} &  & 256 &-  & \checkmark & - & - &8$\times$9.8/8$\times$75.9 & \color{HighlightColor}23.91 & \color{HighlightColor}32.12 & 29.45 & \color{HighlightColor}24.17 & \color{HighlightColor}25.67 & \color{HighlightColor}29.01 & \color{HighlightColor}22.31 & 15.37 & \color{HighlightColor}25.25\\
 		16 & {mono} &  & 256 &\checkmark  & \checkmark & - & - &8$\times$13/8$\times$79.1 & \color{HighlightColor}24.39 & \color{HighlightColor}32.33 & 29.46 & \color{HighlightColor}24.07 & \color{HighlightColor}25.72 & 29.84 & \color{HighlightColor}22.07 & \color{HighlightColor}15.25 & \color{HighlightColor}25.39\\
 		
 		\cmidrule{4-18}
 		17 & {mono} & & - & - &-& - & \checkmark & 8$\times$33.4/8$\times$74.3& 24.95 & 32.85 & 30.33 & 25.02 & 26.08 & 29.97 & 23.01 & 15.69 & 25.99\\
 		
 		18 & {mono} &  & - & - & - & \checkmark & \checkmark & 8$\times$74.3/8$\times$74.3 & 24.77 & 32.35 & 30.14 & \color{HighlightColor}24.79 & \color{HighlightColor}25.79 & 29.85 & \color{HighlightColor}22.71 & 15.77 & 25.77\\
 		\bottomrule
 	\end{tabular}
 }
 \caption{\label{tab:specializing}BLEU on MuST-C dev set for \textbf{specialization}. We refer to Table~\ref{tab:adating_finetuning_ablation} for all notation.%
 }
\end{table*}

In addition to the refinement recipe where language-specific adapters tailor the frozen multilingual ST model to translate in the corresponding direction, we also propose a recipe to facilitate the specialization in individual language pairs: by replacing the multilingual vocabulary by the monolingual ones corresponding to each target language. This recipe allows us to transfer from multilingual models to monolingual ones. A practical benefit is that one can easily leverage pre-trained multilingual models for new languages.

Table~\ref{tab:specializing} displays the results of the specializing recipe. Starting from a trained multilingual ST model, one can obtain an improvement of $1.3$--$1.4$ BLEU on average ({row 8} \vs row 1 and 2) compared to the bilingual and multilingual baselines trained from scratch for the small architecture where $D=256$. However, for a larger network ($D=512$), the gain is more modest ($0.4$ BLEU on average).

\end{document}